\documentclass[lettersize,journal]{IEEEtran}
\usepackage{amsmath,amsfonts}
\usepackage{algorithmic}
\usepackage{algorithm}
\usepackage{array}
\usepackage[caption=false,font=normalsize,labelfont=sf,textfont=sf]{subfig}
\usepackage{textcomp}
\usepackage{stfloats}
\usepackage{url}
\usepackage{verbatim}
\usepackage{graphicx}
\usepackage{cite}
\usepackage{booktabs}
\usepackage{multirow}
\usepackage{xcolor}

\usepackage[switch]{lineno}

\begin{document}

\title{Audio-Visual Segmentation via Depth-Guided Collaborative Modeling}

\author{
Zhaojin Fu, Yuyang Hong, Qi Yang, Zili Wang, Kun Ding, Shiming Xiang, and Bin Fan\textsuperscript{*}
\thanks{This work was supported by the National Natural Science Foundation of China (Grant Nos. U24A20218 and 62306310).}
\thanks{Zhaojin Fu and Bin Fan are with the School of Intelligent Science and Technology, University of Science and Technology Beijing, Beijing 100083, China (e-mail: d202410395@xs.ustb.edu.cn; bin.fan@ieee.org).}
\thanks{Yuyang Hong, Qi Yang, Zili Wang, Kun Ding, and Shiming Xiang are with the Institute of Automation, Chinese Academy of Sciences, Beijing 100190, China. (e-mail: hongyuyang2023@ia.ac.cn; yangqi2021@ia.ac.cn; ziliwang2022@ia.ac.cn; kun.ding@ia.ac.cn; smxiang@nlpr.ia.ac.cn)}
\thanks{\textsuperscript{*}Corresponding author: Bin Fan (e-mail: bin.fan@ieee.org).}
}

\markboth{Journal of \LaTeX\ Class Files,~Vol.~14, No.~8, August~2021}%
{Shell \MakeLowercase{\textit{et al.}}: A Sample Article Using IEEEtran.cls for IEEE Journals}



\maketitle

\begin{abstract}
Audio-Visual Segmentation (AVS) is a fundamental task in multimodal perception that performs pixel-level segmentation of sounding objects in videos by leveraging both visual and audio cues. It has broad applications in video understanding, human–computer interaction, and autonomous driving.
However, most existing AVS methods do not explicitly model geometric cues such as relative distance and occlusion, thereby limiting the robustness of cross-modal alignment. In human perception, spatial structure is naturally integrated with audio-visual evidence to accurately localize sounding objects. Motivated by this, we incorporate estimated depth as a spatial structural cue for AVS and propose DGCM-AVS, a tri-modal framework that jointly models audio, visual, and depth information. Specifically, we design a Depth-Aware Dynamic Modulator to improve the separation of adjacent objects while preserving intra-object feature consistency. Furthermore, we propose Depth-Guided Progressive Fusion, which uses depth as an intermediate bridge to progressively align audio cues with visual features. Compared to state-of-the-art methods, DGCM-AVS achieves relative improvements of 10.2\% in \(\mathcal{M}_\mathcal{J}\) and 8.7\% in \(\mathcal{M}_\mathcal{F}\) on the AVSS dataset. We believe our study highlights depth as a promising yet underexplored modality for AVS and may encourage further research in this direction.

\end{abstract}


\section{Introduction}

\IEEEPARstart{W}{ith} auditory cues, humans can focus visual attention on sounding objects. This cross-modal perceptual mechanism is crucial for understanding complex scenes. As applications such as human-computer interaction and autonomous driving continue to expand, it becomes increasingly important to endow machines with similar cross-modal perceptual capabilities. Accordingly, Audio-Visual Segmentation (AVS) \cite{11, 14, liu2025robust} has attracted increasing attention in recent years. In its general form, AVS leverages both audio and visual cues to localize sounding objects and segment them at the pixel level. Depending on the setting, AVS can be formulated as binary foreground-background segmentation, or as a semantic task that additionally predicts category labels for the segmented sounding objects.


\begin{figure}[!t]
  \centering
  \includegraphics[width=1\linewidth]{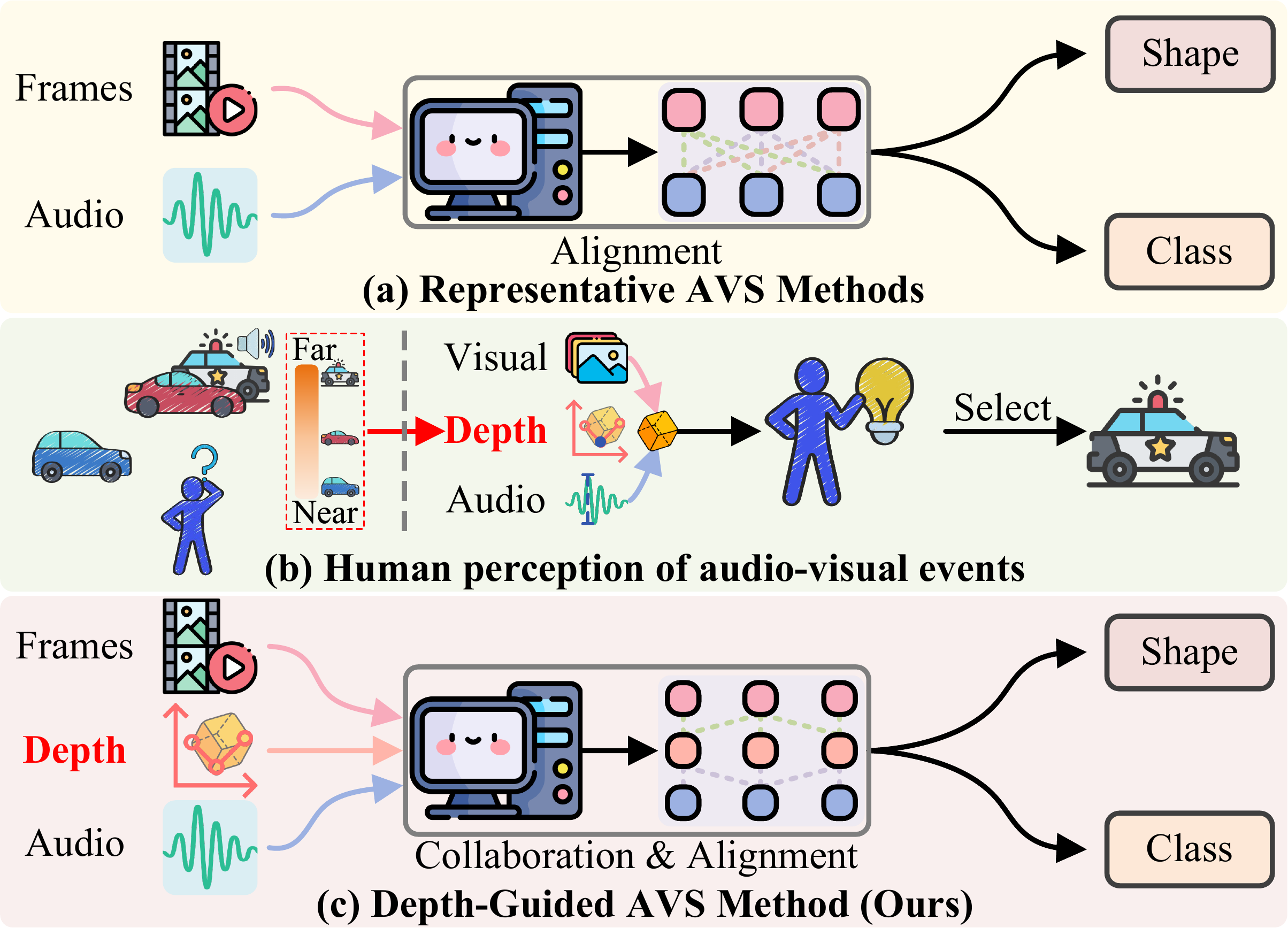}
  \caption{
  (a) Representative AVS methods that align audio and visual features.
  (b) Illustrates that humans exploit depth cues together with audio evidence to disambiguate sounding objects.
  (c) Presents our core idea, which aims to perform the AVS task in a human-like way.
  }
  \label{fig:1}
  \vspace{-0.4cm}
\end{figure}

Existing AVS methods can be broadly viewed as improving either cross-modal semantic alignment or mask boundary quality, although many designs combine these ingredients in different forms. One representative line focuses on semantic alignment \cite{14, 17, 21}, as shown in Fig. \ref{fig:1}(a), where audio is used to modulate semantic visual features through cross-modal interaction. However, semantic cues alone are often insufficient for precise localization and boundary delineation, especially when multiple candidates have similar appearances or the sounding object is partially occluded. Another line leverages strong visual priors from large-scale vision models to refine segmentation masks \cite{wang2025sam2, 39}. With visual prompting, these methods can produce sharper boundaries and improved region separation. Nonetheless, excessive prompt-induced partitioning can over-segment spatially adjacent regions, undermining object integrity and causing fragmented masks or semantic inconsistency. Despite their progress, the lack of explicit spatial-structure constraints can makes cross-modal correspondence less robust. As a result, segmentation can become unstable in complex scenes, often exhibiting attention drift to salient but silent regions and boundary leakage.

Inspired by human spatial perception, we suggest that explicitly introducing scene spatial structure into AVS can constrain cross-modal correspondence, thereby improving the stability and robustness of segmentation. In particular, when multiple candidate objects have similar appearances or are partially occluded, spatial cues such as depth ordering and occlusion relationships can help determine which visible entity is more likely to correspond to the audio signal. This reduces attention drift and boundary leakage, as illustrated in Fig. \ref{fig:1}(b). Recently, researchers have increasingly recognized that spatial information is crucial for human-like multimodal perception and have begun exploring its integration into sound source localization \cite{32} and multimodal learning \cite{zhu2024languagebind}. However, these efforts do not explicitly model the pixel-level localization and fine boundary delineation required by AVS, and therefore cannot directly satisfy the requirements for accurate segmentation. Therefore, a more systematic approach is needed to endow AVS models with spatial awareness, enabling more robust cross-modal alignment.

We propose the \textbf{\underline{D}}epth-\textbf{\underline{G}}uided \textbf{\underline{C}}ollaborative \textbf{\underline{M}}odeling for \textbf{\underline{A}}udio-\textbf{\underline{V}}isual \textbf{\underline{S}}egmentation (DGCM-AVS) to foster more effective collaboration among the audio, visual, and depth modalities. Since depth provides a direct cue to scene spatial structure, we incorporate it into our model design, as shown in Fig. \ref{fig:1}(c), to equip the model with two capabilities: (1) improve the discrimination of objects at different distances and positions while preserving structural integrity. (2) using depth as a bridge to align audio and visual features more effectively.

For capability (1), we propose the Depth-Aware Dynamic Modulator (DADM), which adopts a scale-based approximation of frequency separation. Specifically, it uses neighborhood aggregation to form a stable, region-level low-frequency visual representation, and leverages a depth residual to highlight boundary-sensitive high-frequency structural cues. By fusing these complementary cues, DADM improves inter-object discrimination while preserving intra-object consistency.
For capability (2), we propose Depth-Guided Progressive Fusion module (DGPF), which uses depth information as a bridge to align audio and visual features in two stages. In the first stage, Audio Depth Symbiotic, audio features are projected into the geometric space to establish the correspondence between sound and scene structure. In the second stage, Target Search, audio and visual semantics are aligned in the spatial domain to produce accurate segmentation masks.

In this paper, our main contributions are as follows:
\begin{itemize}
\item To the best of our knowledge, DGCM-AVS is the first to explore the potential of using depth information in AVS. It introduces a collaborative modeling framework across audio, visual, and depth information. This method offers new insights for advancing the AVS field and achieves state-of-the-art performance on both the AVSBench-Object and AVSBench-Semantic.
\item We propose the Depth-Aware Dynamic Modulator module to embed depth information into image features. This design enhances intra-object consistency and strengthens inter-object discrimination.
\item We propose the Depth-Guided Progressive Fusion module for gradual interaction across audio, visual, and depth cues. Depth information serves as a bridge to improve alignment between audio and visual features.
\end{itemize}

\section{Related Work}

\subsection{Audio-Visual Segmentation}

AVS aims to segment sounding objects at the pixel level by jointly leveraging audio and visual cues in complex scenes. AVSBench \cite{17} introduced two benchmark datasets, AVSBench-Object and AVSBench-Semantic, and proposed a baseline that embeds audio features into visual representations for sounding-object localization. Specifically, AVSBench-Object focuses on binary foreground–background segmentation. AVSBench-Semantic defines a more challenging setting where models are required to segment sounding objects and predict their semantic categories. Following this, several studies \cite{43,50,68,62,26,11,mao2023multimodal} approached the task from the perspective of semantic alignment, using audio cues as queries to retrieve visual targets and enhance cross-modal representations. ECMVAE \cite{mao2023multimodal} formulates audio-visual segmentation as a conditional multimodal variational autoencoding problem. It disentangles modality-shared and modality-specific latent codes, and optimizes them with orthogonality constraints and cross-modal mutual-information regularization to better capture sounding objects. While effective in establishing semantic-level correspondence, semantic-alignment-based methods generally lack explicit spatial constraints and thus struggle with precise localization in complex scenes. Another line of work \cite{wang2025sam2, 39} explored boundary enhancement with large vision models, for example using SAM \cite{22} to provide visual prompts. These approaches improved boundary sharpness but often over-emphasized local edges, which compromised object integrity. Other studies \cite{24,40} introduced textual descriptions to support audio-visual alignment. Although textual guidance enriches cross-modal connections, it can be unreliable for target localization under distractors, occlusions, or cluttered scenes. In summary, existing methods have explored semantic alignment, boundary enhancement, and semantic description, but none explicitly model the spatial structure of scenes. To address this limitation, we propose DGCM-AVS, which introduces depth as explicit geometric cue to strengthen cross-modal correspondence and improve the consistency and robustness of AVS predictions.

\vspace{-0.2cm} 
\subsection{Depth estimation}

Depth estimation predicts a pixel-wise depth map from a monocular or stereo image to describe the 3D geometry of a scene. Recently, some methods learn depth with explicit geometric constraints and aim for metric-consistent structure \cite{61, hu2024metric3d, 37}. Other methods \cite{chen2025video, 34, 35} emphasize large-scale pretraining to learn transferable depth representations and achieve strong generalization. Although their designs differ, they often provide reliable cues about relative layout, depth gradients, and depth discontinuities in practice. 

In computer vision \cite{torres2024davide, sun2024depth, cheng2020zero, guo2025compressed, guo2025depth, liu2025segmenting, liu2025multi}, depth maps are commonly adopted as transferable geometric priors, and they are particularly useful under appearance degradation such as low illumination, blur, and occlusion, where RGB features are more affected by texture and lighting changes. In deblurring, DAVIDE \cite{torres2024davide} injects depth features into the network across multiple stages and uses depth-aware fusion. The gains are larger when the temporal context is short, suggesting that depth helps stabilize structure recovery when appearance evidence is weak. In semantic segmentation, Texture-Diffusion \cite{sun2024depth} shows that depth provides reliable layout and boundary discontinuity cues. It uses cross-modal diffusion and structural consistency constraints to reduce information loss caused by modality gaps, which improves robustness. For low-light segmentation, DPSAM \cite{liu2025segmenting} introduces depth into feature refinement via cross-modal attention. This reduces failures caused by degraded RGB features and texture confusion, and it improves prediction stability and boundary quality. For occluded overlapped text segmentation, MOTS \cite{liu2025multi} uses depth-guided decoding and cross-attention to model occlusion hierarchies, which improves boundary localization in overlapped scenes.

Overall, these studies show that depth information often provides a more stable global layout and scale structure, which supports region-level localization. It also offers clearer boundary discontinuities and occlusion breaks, which improves boundary adherence and detail recovery. Since these geometric cues are usually more robust than RGB texture and illumination, depth can provide structural constraints for cross-modal representation learning and can improve cross-modal alignment to some extent \cite{32,zhu2024languagebind}. These insights also match key challenges in audio-visual segmentation. Audio cues are often coarse and region-level, and occlusion can more easily cause candidate confusion and boundary drift. Based on this, we introduce estimated depth into DGCM-AVS as an auxiliary geometric prior. We focus on boundary and occlusion discontinuities and global layout cues. During feature fusion, these cues provide more stable structural constraints for audio-guided localization and further refine segmentation boundaries, which improves audio-visual alignment and final segmentation performance in occluded scenes.


\begin{figure*}
  \centering
  \includegraphics[width=1\linewidth]{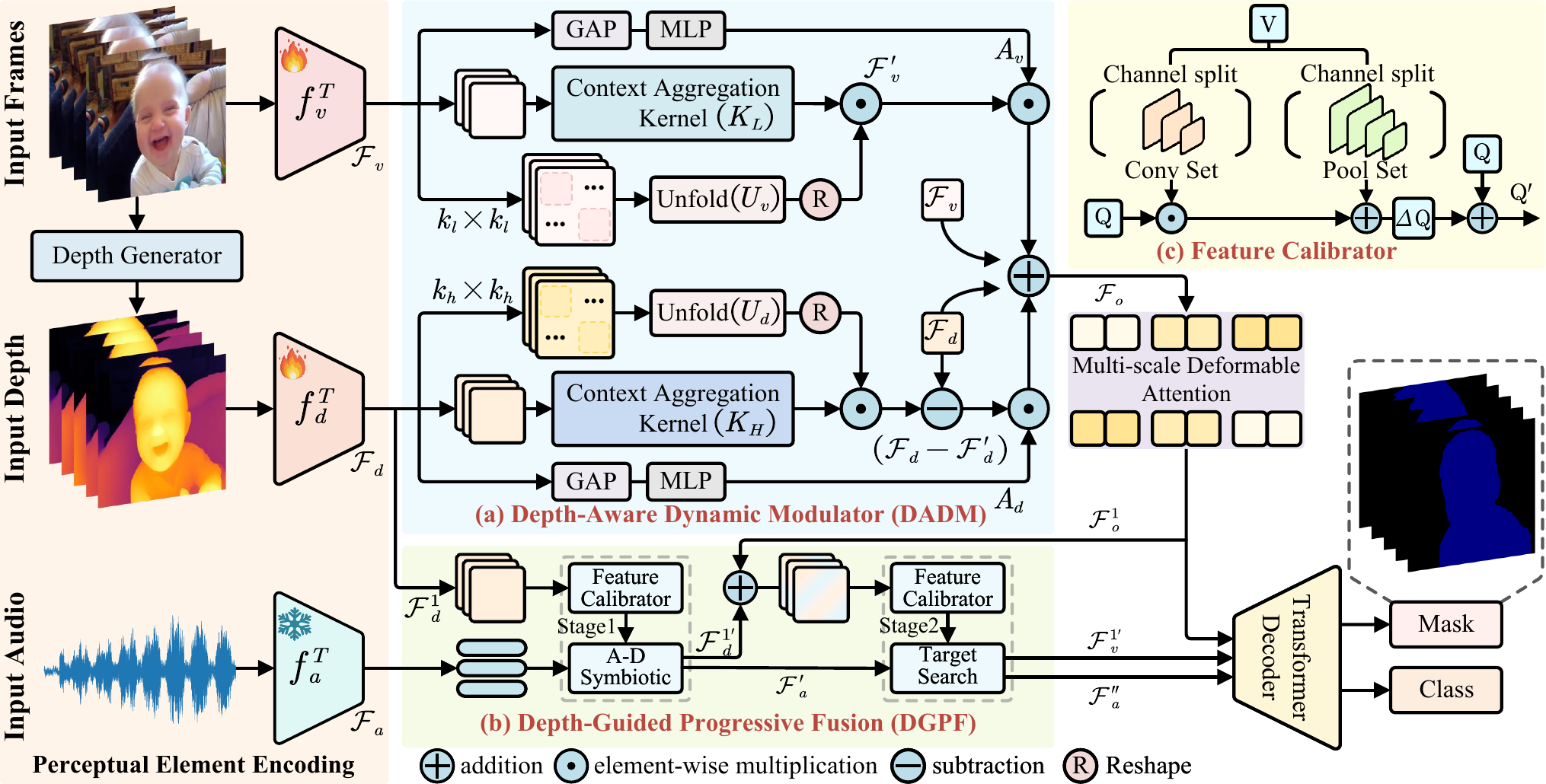}
  \caption{Overview of the proposed DGCM-AVS. (a) DADM extracts different frequency components from image and depth features to improve intra-object consistency and enhance inter-object distinction. (b) DGPF uses depth information as a bridge to progressively complete the Audio Depth Symbiotic stage (A-D Symbiotic) and Target Search stages, achieving cross-modal feature alignment. (c) Feature Calibrator further strengthens the target-related feature responses.}
  \label{fig:2}
\end{figure*}

\begin{algorithm}[tb]
\caption{DGCM-AVS for AVSS}

\textbf{Input}: Train dataset \(\mathcal{D}\), mask label \(\mathcal{M}\), class label \(\mathcal{C}\)\\
\textbf{Output}: \(\mathcal{M}_{out}\), \(\mathcal{C}_{out}\) 
\begin{algorithmic}[1] 
\STATE The total loss \( \mathcal{L}=\lambda_{bce}\mathcal{L}_{bce}+\lambda_{dice}\mathcal{L}_{dice}+\lambda_{ce}\mathcal{L}_{ce} \), with \( \lambda_{bce} = 5 \), \( \lambda_{dice} = 5 \), \( \lambda_{ce} = 2 \)
\FOR{\( x_v,x_{mel}\in \mathcal{D}\)}
    \STATE \(\mathcal{F}_v=f_v^T(x_v)\), \(\mathcal{F}_v = \{ \mathcal{F}_{v}^{i} \}_{i=1}^{4}\)
    \STATE Depth information: \(x_d= \text{Depth Generator}(x_v)\)
    \STATE \(\mathcal{F}_d=f_d^T(x_d)\), \(\mathcal{F}_d = \{ \mathcal{F}_{d}^{i} \}_{i=1}^{4}\)
    \STATE \(\mathcal{F}_a=f_a^T(x_{mel})\)
    \STATE Depth-Aware Dynamic Modulator \(f_{DADM}(\cdot)\):
    \FOR{\(i\) in \(4\)}
        \STATE \(\mathcal{F}_{o}^{i}=f_{DADM}^{i}(\mathcal{F}_{v}^{i},~\mathcal{F}_{d}^{i})\)
    \ENDFOR
    \STATE Multi-scale Deformabl Attention \(f_{MSDA}(\cdot)\):
    \STATE \hspace{1em}\(\mathcal{F}_{o}=f_{MSDA}(\mathcal{F}_{o})\)
    \STATE Depth-Guided Progressive Fusion \(f_{DGPF}(\cdot)\):
    \STATE \hspace{1em}\( \mathcal{F}_{v}^{1'},\mathcal{F}_{a}^{''}=f_{DGPF}(\mathcal{F}_{a},~\mathcal{F}_{d},~\mathcal{F}_{o})\)
    \STATE Transformer Decoder \(f_{Deocder}(\cdot)\):
    \STATE \hspace{1em}\(\mathcal{M}_{out},~\mathcal{C}_{out}= f_{Deocder}(~\mathcal{F}_{a}^{''},~\mathcal{F}_v^{1'},~\mathcal{F}_o)\)
    \STATE \( Loss=\mathcal{L}((\mathcal{C}_{out},~\mathcal{C}),~(\mathcal{M}_{out},~\mathcal{M}))\)
\ENDFOR
\STATE \textbf{return} \(Loss\)
\end{algorithmic}
\label{alg:algorithm}
\end{algorithm}

\section{Method}

\subsection{Overall Architecture}

The DGCM-AVS framework is shown in Fig. \ref{fig:2}. The Perceptual Element Encoding first builds perceptual representations from video, depth, and audio. The Depth-Aware Dynamic Modulator injects depth-derived structural cues into visual features via a dynamic low-/high-frequency fusion scheme. This enhances the model’s ability to discriminate adjacent objects, improving inter-object separability. Subsequently, to capture object details across different scales, we apply Multi-Scale Deformable Attention, inspired by Deformable DETR \cite{zhu2020deformable}, enabling effective multi-scale feature interaction. The Depth-Guided Progressive Fusion is built upon depth information and adopts a two-stage strategy to progressively integrate audio and visual features for audio-visual alignment. This design is inspired by how humans identify sounding objects in complex environments. Finally, we adopt the Transformer decoder \cite{52} to reconstruct the final segmentation masks. To further explain the proposed DGCM-AVS, we present its algorithmic process in Algorithm~\ref{alg:algorithm}. The algorithm shows the complete pipeline from multi-modal feature extraction to semantic segmentation.

\vspace{-0.2cm} 
\subsection{Perceptual Element Encoding} 
\label{3.2}
Given an input video \(x_v \in \mathbb{R}^{T \times 3 \times H \times W}\), we first extract visual features using a visual encoder \(f_v^T(\cdot)\), producing multi-scale features \(\mathcal{F}_v = \{ \mathcal{F}_{v}^{i} \}_{i=1}^{4}\), where \(\mathcal{F}_{v}^{i} \in \mathbb{R}^{T \times C_i \times \frac{H}{2^{i+1}} \times \frac{W}{2^{i+1}}}\). Since current AVS benchmarks do not include depth information, we use Depth Anything V2 \cite{35} to estimate depth information \(x_d\), which is converted to RGB format to match the visual input. A separate depth encoder \(f_d^T(\cdot)\) is used to extract multi-scale depth features \(\mathcal{F}_d = \{ \mathcal{F}_{d}^{i} \}_{i=1}^{4}\), with \(\mathcal{F}_{d}^{i} \in \mathbb{R}^{T \times C_i \times \frac{H}{2^{i+1}} \times \frac{W}{2^{i+1}}}\). For both the visual encoder \(f_v^T(\cdot)\) and the depth encoder \(f_d^T(\cdot)\), we experiment with two backbone networks: ResNet-50 \cite{49} and PVT-v2 \cite{19}. For audio, the raw waveform is first resampled to a 16 kHz mono signal. We then compute the Mel-spectrogram \(x_{mel} \in \mathbb{R}^{T \times 96 \times 64}\) using Short-Time Fourier Transform. A pretrained VGGish \(f_a^T(\cdot)\) \cite{51} is used to extract frame-level audio features \(\mathcal{F}_a \in \mathbb{R}^{T \times D}\), where \(D = 128\) and \(T\) is the number of video frames.

\subsection{Depth-Aware Dynamic Modulator}
\label{3.3}

Visual features contain both low- and high-frequency components. The low-frequency part mainly describes global structure and semantic layout, providing stable and robust cues for coarse localization. In contrast, the high-frequency part contains richer fine textures, but it is more sensitive to noise and occlusion, and it often struggles to distinguish visually similar objects, which may confuse the model. Depth provides complementary benefits. In the spatial domain, depth corresponds more directly to geometry and discontinuities, so it can produce cleaner local changes around boundaries and occlusions. For AVS, audio often provides region-level cues related to the sounding object. Therefore, we need stable semantics and layout to support audio-visual alignment, and we also need reliable boundary cues to refine the mask.

\begin{figure}
  \centering
  \includegraphics[width=.96\linewidth]{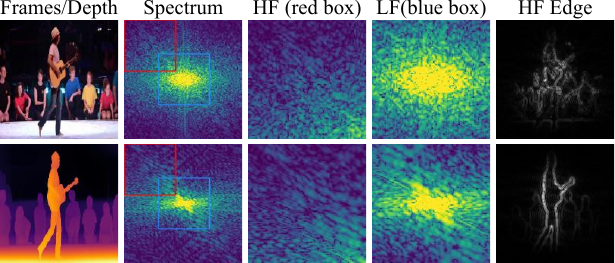}
  \caption{Frequency-domain comparison of RGB and depth. The figure shows log power spectra, zoomed views of low/high-frequency regions, and edge responses reconstructed from the mid-to-high frequency band.}
  \label{fig:fft}
  \vspace{-0.4cm}
\end{figure}

As shown in Fig. \ref{fig:fft}, we visualize the RGB frame and its depth map in the frequency domain. Both spectra show strong energy near the center, which suggests that the two modalities share global structural information. However, their behaviors differ in the high-frequency region. The RGB spectrum is more scattered and is easily affected by textures and lighting changes. These appearance-driven components can introduce sound-irrelevant distractions in AVS and reduce the stability of audio-visual alignment. In contrast, the depth spectrum is more concentrated along contours and shows clearer structure with less interference. The reconstructed high-frequency edges are also more continuous and align better with object boundaries. Based on these observations, we use low-frequency visual cues as a stable semantic basis for audio-guided coarse localization. We use high-frequency depth cues to tighten and refine the boundaries.

Motivated by the above observations, we propose the Depth-Aware Dynamic Modulator (DADM) in Fig. \ref{fig:2} (a). To provide an explicit and reproducible definition of high- and low-frequency components, we adopt a scale-based approximation for frequency separation in the spatial domain. Neighborhood aggregation is treated as a low-pass operator, since aggregating features over a larger receptive field performs stronger local averaging, suppresses rapid spatial variations, and yields a low-frequency representation. Under this formulation, the low-frequency component is the aggregated feature, whereas the high-frequency component is the residual between the original feature and the aggregated feature, \(\text{HighFreq}(F)=F-\text{LowFreq}(F)\). This residual emphasizes fast-changing structural cues in depth, including boundary discontinuities and occlusion-induced breaks. Following this design, the low-frequency branch uses a larger window \(k_l=5\) to integrate wider context and produce a stable semantic basis that aligns with region-level audio guidance and preserves global layout consistency. The high-frequency branch uses a smaller window \(k_h=3\) to retain finer local variations, which enables the residual signal to capture depth discontinuities at object boundaries more effectively.

Formally, the visual features \( \mathcal{F}_v \) and depth features \( \mathcal{F}_d \) are first processed with channel attention to generate importance weights \(A_v\) and \(A_d\), which emphasize feature channels related to sounding objects. Then, dynamic perception kernels are constructed along high- and low-frequency paths.

For the high-frequency path, we first obtain compressed features \( \tilde{\mathcal{F}}_d=\text{Conv}_{1\times 1}\left( \mathcal{F}_d \right) \in \mathbb{R}^{T\times c\times h\times w}\), and we then process it within a \(k_h=3\) window to focus on local depth variations. A kernel normalization operation generates a local perception kernel \(K_H\):
\begin{equation}
     K_H=\mathcal{T}\left( \phi \left( \text{HConv}\left( \tilde{\mathcal{F}}_d \right) \right) \right) \in \mathbb{R}^{T\times 1\times k_h^2\times h\times w},
    \label{eq:1} 
\end{equation}
The perception kernel is constructed based on the input features, allowing it to adapt dynamically to different inputs and enabling input-dependent feature perception. Here, \(\mathcal{T}\left( \cdot \right)\) represents the kernel normalization operation \(\sum_{p=1}^{k_h^2}{K_{H}^{\left( p \right)}=1}\), \(\phi \left( \cdot \right) \) represents a feature reshaping function into format \(T\times1\times k_h^2\times h\times w\), and \(\text{HConv}\left( \cdot \right) \) denotes a convolutional operator with kernel size \(3\times 3\).

Moreover, the feature \(\tilde{\mathcal{F}}_d\) is padded using reflection padding to obtain \(\tilde{\mathcal{F}}_d'\in \mathbb{R}^{T\times c\times \left( h+2p \right) \times \left( w+2p \right)}\), \(p=\left\lfloor k_h/2\right\rfloor\), \(k\in \left\{ 1,...,k_{h}^{2} \right\} \) represents the linear index of a pixel within the \(k_h\times k_h\) neighborhood. For each spatial location \( (i, j) \), a \( k_h \times k_h \) receptive window is centered to extract neighboring features. These are then unfolded and reshaped to obtain the aggregated neighborhood feature \( U_d \):
\begin{equation}
     U_d = \text{Unfold}(\tilde{\mathcal{F}}_d') \in \mathbb{R}^{T \times c \times k_h^2 \times h \times w},
    \label{eq:2} 
\end{equation}
This design allows the model to capture relative spatial relationships among neighboring pixels while maintaining spatial sensitivity during subsequent high-frequency kernel computations. Similarly, in the low-frequency perception branch, a \( k_l=5 \) receptive window is applied to compute the low-frequency kernel \(K_L\in\mathbb{R}^{T\times1\times{k_l}^2\times h\times w} \) and the corresponding aggregated neighborhood feature \( U_v\in\mathbb{R}^{T\times c\times{k_l}^2\times h\times w} \). 

Finally, \(U_d\) and \(U_v\) are convolved with their respective perception kernels \(K_H\) and \(K_L\). Channel attention and residual fusion are applied to jointly leverage region-level semantics and boundary-sensitive depth cues. This process can be formulated as:
\begin{equation}
    \mathcal{F}_d'=\sum_{k=1}^{k_{h}^{2}}{\left( U_d\left[ :,:,k,:,: \right] \odot K_H\left[ :,0,k,:,: \right] \right)}, 
    \label{eq:3} 
\end{equation}
\begin{equation}
    \mathcal{F}_v'=\sum_{k=1}^{k_{l}^{2}}{\left( U_v\left[ :,:,k,:,: \right] \odot K_L\left[ :,0,k,:,: \right] \right)},
    \label{eq:4} 
\end{equation}
\begin{equation}
    \mathcal{F}_{d}^{H}=\left( \mathcal{F}_d-\mathcal{F}_d'\right) \odot A_d+\mathcal{F}_d , \mathcal{F}_{v}^{L}=\mathcal{F}_v'\odot A_v+\mathcal{F}_v ,
    \label{eq:5} 
\end{equation}
\begin{equation}
    \mathcal{F}_o=\text{Conv}_{3\times 3}\left( \mathcal{F}_{d}^{H}+\mathcal{F}_{v}^{L} \right),
    \label{eq:6} 
\end{equation}
where \(\odot\) represents element-wise multiplication, \(\mathcal{F}_o\in \left\{ \mathcal{F}_{o}^{i} \right\} _{i=1}^{4}\), \(\mathcal{F}_{o}^{i}\in \mathbb{R}^{T\times C_i\times \frac{H}{2^{i+1}}\times \frac{W}{2^{i+1}}}\). \(\mathcal{F}_d'\) can be interpreted as a locally aggregated approximation of the depth feature that suppresses rapid spatial variations. The high-frequency depth cue is explicitly represented by the residual \(\mathcal{F}_d-\mathcal{F}_d'\), which exhibits strong responses at depth discontinuities and object boundaries. This residual is then reweighted by \(A_d\) and added back to \(\mathcal{F}_d\) to enhance boundary sensitivity. Meanwhile, \(\mathcal{F}_v'\) provides a smoother, low-frequency visual representation that is less affected by texture noise, supporting stable AVS.

\begin{figure}
  \centering
  \includegraphics[width=.96\linewidth]{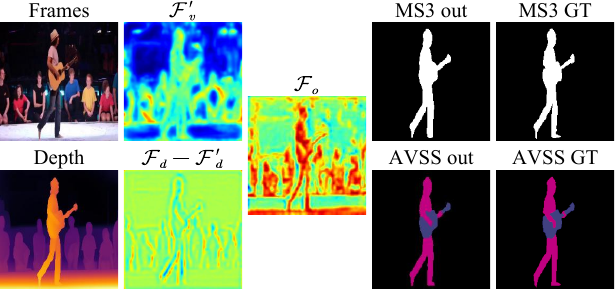}
  \caption{Visualization of Intermediate Features in DADM. We visualize the intermediate features from the low-frequency branch \(\mathcal{F}_v'\), the high-frequency branch \(\mathcal{F}_d-\mathcal{F}_d'\), and the final fused feature \(\mathcal{F}_o\). We also show the model predictions on the MS3 and AVSS benchmarks, together with the corresponding GT.}
  \label{fig:hot_map}
  \vspace{-0.4cm}
\end{figure}

As shown in Fig.~\ref{fig:hot_map}, we visualize the two branches of DADM and their fused feature. The low-frequency branch output \(\mathcal{F}_v'\) mainly shows region-level responses. It can stably cover the main area of the sounding object and reflects strong semantic consistency. However, its responses are often smooth, which may lead to blurred boundaries or slight over-expansion. In contrast, the high-frequency branch \(\mathcal{F}_d-\mathcal{F}_d'\), constructed from depth residuals, responds more strongly to depth discontinuities. It highlights object contours and local structural changes, providing cleaner geometric cues for boundary refinement. The fused feature \(\mathcal{F}_o\) effectively combines these complementary cues. It preserves the integrity of the object region while enhancing contour clarity and suppressing background interference. Accordingly, the predictions on the MS3 and AVSS benchmarks match the GT more closely, especially along boundaries and in local details, which further verifies the effectiveness of DADM.

\vspace{-0.2cm} 
\subsection{Depth-Guided Progressive Fusion}
\label{3.4}

\begin{figure}
  \centering
  \includegraphics[width=.94\linewidth]{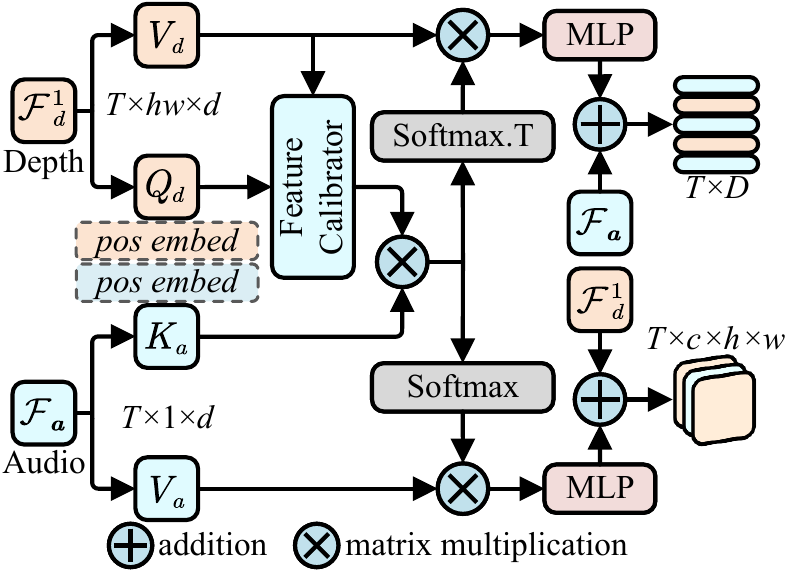}
  \caption{Depth Bridge for DGPF. Serving as a structural cue, depth embeds features in A–D symbiosis and facilitates audio–visual alignment during Target Search.}
  \label{fig:3}
  \vspace{-0.4cm}
\end{figure}

Motivated by the observation that depth provides explicit spatial cues, we incorporate it into the model to enhance audio-visual alignment. To this end, we propose the Depth-Guided Progressive Fusion (DGPF) (Fig. \ref{fig:2}(b)), which leverages depth as a bridge to progressively establish semantic correspondence between audio and sounding objects. It consists of two stages: in Audio Depth Symbiotic, audio features are projected into the spatial domain to highlight candidate sounding regions; in Target Search, depth guidance together with audio-visual semantics enables precise localization and segmentation. A Depth Bridge is designed for both stages to enable effective fusion, as in Fig. \ref{fig:3}. Unlike conventional fusion strategies, DGPF explicitly exploits depth as a progressive bridge, ensuring geometric alignment and semantic consistency.

In the Audio Depth Symbiotic stage, the depth feature \(\mathcal{F}_{d}^{1}\) is first encoded with sine-based positional encoding to generate the query vector \(Q_d = \mathcal{F}_{d}^{1p} W_Q\). In parallel, the value vector \(V_d = \mathcal{F}_{d}^{1} W_V^d\) is obtained directly from \(\mathcal{F}_{d}^{1}\) through a linear transformation. Then, the Feature Calibrator performs multi-scale processing on \(V_d\), applies query-guided weighting, and incorporates the result into \(Q_d\), yielding the updated query \(Q_d'\). The audio value vector is computed as \(V_a = \mathcal{F}_a W_V^a\), and the key is obtained by adding learnable positional encoding to the audio features, resulting in \(K_a = \mathcal{F}_a^p W_K\). By sharing \(Q_d'\) and \(K_a\), the model projects audio features into the spatial domain and highlights candidate sounding regions more effectively. This design follows a common practice in Transformer \cite{20} and DETR \cite{zhu2020deformable}. Injecting positional information into the query \(Q\) and the key \(K\) is sufficient to make the attention weights spatially aware, while the value \(V\) mainly preserves the semantic content to be aggregated. This process can be formulated as:

\begin{equation}
    \mathcal{F}_{d}^{1'}=\text{Softmax} \left( \frac{Q_d'K_{a}^{T}}{\sqrt{d}} \right) V_a+\mathcal{F}_{d}^{1}, 
    \label{eq:7} 
\end{equation}
\begin{equation}
    \mathcal{F}_{ad}=\text{Softmax} \left(\left( \frac{Q_d'K_{a}^{T}}{\sqrt{d}} \right)^{T}\right) V_d+\mathcal{F}_a .
    \label{eq:8} 
\end{equation}

We propose the Feature Calibrator (Fig. \ref{fig:2}(c)), which adopts a dual-branch design. One branch applies multi-scale depthwise separable convolutions \(\mathcal{C}_k(\cdot)\) to capture boundary textures, while the other uses adaptive pooling \(\mathcal{P}_s(\cdot)\) followed by bilinear interpolation \(I_s(\cdot)\) to model global spatial layout. The query \(Q_d\) modulates the texture features to suppress irrelevant noise, and the spatial layout features provide a stable structure for alignment, thereby enhancing intra-object consistency. The updated query is computed as \(\Delta Q = Q_d \odot \text{Concat}\left( \mathcal{C}_k(V_d) \right)_{k = \{3,5,7\}} + \text{Concat}\left( I_s( \mathcal{P}_s(V_d) ) \right)_{s = \{1,2,3,6\}}\), and the final query becomes \(Q_d' = Q_d + \Delta Q\). The Feature Calibrator enhances the discriminability of target-related features beyond conventional query-modulation strategies.

Next, we perform modality-level aggregation: \(\mathcal{F}_{v}^{1} = \mathcal{F}_{d}^{1'} + \mathcal{F}_{o}^{1}\), \(\mathcal{F}_a' = \mathcal{F}_{ad} + \mathcal{F}_a\). Here, the depth feature \(\mathcal{F}_{d}^{1'}\) already contains audio information, while the visual feature \(\mathcal{F}_{o}^{1}\), fused by DADM, is rich in semantic content and encodes scene geometry. Similarly, the updated audio feature \(\mathcal{F}_a'\) also contains depth information. 

Finally, in the Target Search stage, we apply Depth Bridge to \(\mathcal{F}_{v}^{1}\) and \(\mathcal{F}_a'\), using depth information as a bridge to improve alignment between the audio and visual modalities. Unlike the first stage, which focuses on candidate region enhancement, Target Search refines the correspondence by jointly leveraging audio, visual, and depth semantics for precise localization and segmentation. The computation follows Equations \eqref{eq:7} and \eqref{eq:8}, resulting in the updated features \(\mathcal{F}_{v}^{1'}\) and \(\mathcal{F}_a^{''}\).

\begin{table*}[t]
\vspace{-0.4cm}
    \renewcommand\arraystretch{1.1} 
    \small
    \centering
    \caption{Quantitative results on the Single Sound Source Segmentation (S4) and Multiple Sound Source Segmentation (MS3), and AVSBench-Semantic (AVSS). Bold indicates the best result, and underline indicates the second best.}
    \setlength{\tabcolsep}{6.3mm} 
    {
        \begin{tabular}{llcccccc}
        \toprule
                                               &                             & \multicolumn{2}{c}{S4}                                                      & \multicolumn{2}{c}{MS3}                                                     & \multicolumn{2}{c}{AVSS}    \\ \cmidrule(r){3-4} \cmidrule(r){5-6} \cmidrule(r){7-8}
            \multirow{-2.3}{*}{Method}               & \multirow{-2.3}{*}{Backbone}  & \(\mathcal{M}_\mathcal{J}\)                                  & \(\mathcal{M}_\mathcal{F}\)                                   & \(\mathcal{M}_\mathcal{J}\)                                  & \(\mathcal{M}_\mathcal{F}\)                                  & \(\mathcal{M}_\mathcal{J}\)                                  & \(\mathcal{M}_\mathcal{F}\)                                   \\ \midrule
        AVSBench \cite{17}                & ResNet-50         & 72.8           & 84.8        & 47.9         & 57.8                                 & 20.2                                 & 25.2                                 \\
        AVS-BiGen \cite{59}               & ResNet-50         & 74.1           & 85.4        & 45.0         & 56.8                                 & -                                    & -                                    \\
        BAVS \cite{53}                    & ResNet-50         & 78.0           & 85.3        & 50.2         & 62.4                                 & 24.7                                 & 29.6                                 \\
        AVSegFormer \cite{21}             & ResNet-50         & 76.4           & 86.7        & 53.8         & 65.6                                 & 26.6                                 & 31.5                                 \\
        SelM \cite{62}                    & ResNet-50         & 76.6           & 86.2        & 54.5         & 65.6                                 & 31.9                                 & 37.2                                 \\ 
        UFE \cite{liu2024audio}           & ResNet-50         & 79.0	       & 87.5        & 55.9         & 64.5                                 & -                                    & -                                    \\
        QDFormer \cite{li2024qdformer}	  & ResNet-50         & 77.6	       & 86.0        & \underline{59.6}	        & 63.5                                 & -	                                  & -                                    \\
        COMBO \cite{39}                   & ResNet-50         & \underline{81.7}           & \underline{90.1}        & 54.5         & \underline{66.6}                                 & \underline{33.3}                                 & \underline{37.3}                                 \\ 
        DiffusionAVS \cite{60}            & ResNet-50         & 75.8           & 86.9        & 49.8         & 62.1                                 & -                                    & -                                    \\ \midrule
        DGCM-AVS (Ours) & ResNet-50 & \textbf{83.3} & \textbf{91.3} & \textbf{61.2} & \textbf{74.4} & \textbf{39.6} & \textbf{43.5} \\ \midrule
        
        AVSBench \cite{17}                & PVT-v2            & 78.7           & 87.9        & 54.0         & 64.5                                 & 29.8                                 & 35.3                                 \\
        AVS-BiGen \cite{59}               & PVT-v2            & 81.7           & 90.4        & 55.1         & 66.8                                 & -                                    & -                                    \\
        BAVS \cite{53}                    & PVT-v2            & 82.0           & 88.6        & 58.6         & 65.5                                 & 32.6                                 & 36.4                                 \\
        AVSegFormer \cite{21}             & PVT-v2            & 83.1           & 90.5        & 61.3         & \underline{73.0}                                 & 37.3                                 & 42.8                                 \\
        SelM \cite{62}                    & PVT-v2            & 83.5           & 91.2        & 60.3         & 71.3                                 & 41.3                                 & \underline{46.9}                                 \\
        UFE	\cite{liu2024audio}           & PVT-v2	          & 83.2           & 90.4        & \underline{62.0}         & 70.9                                 & -	                                  & -                                    \\
        QDFormer \cite{li2024qdformer}	  & Swin-Tiny	      & 79.5	       & 88.2	     & 61.9         & 66.1                                 & -	                                  & -                                    \\
        COMBO \cite{39}                   & PVT-v2            & \underline{84.7}           & \underline{91.9}        & 59.2         & 71.2                                 & \underline{42.1}                                 & 46.1                                 \\ 
        DiffusionAVS \cite{60}            & PVT-v2            & 81.5           & 90.3        & 59.6         & 71.2                                 & 38.1                                    & 43.0                                    \\ \midrule
        DGCM-AVS (Ours)                    & PVT-v2            & \textbf{85.2}  & \textbf{92.4} & \textbf{63.2} & \textbf{75.5} & \textbf{46.4} & \textbf{51.0} \\
        \bottomrule
        \end{tabular}
    }
    \label{tab:1}
    \vspace{-0.4cm}
\end{table*}

\begin{figure*}[h]
  \centering
  \includegraphics[width=1\linewidth]{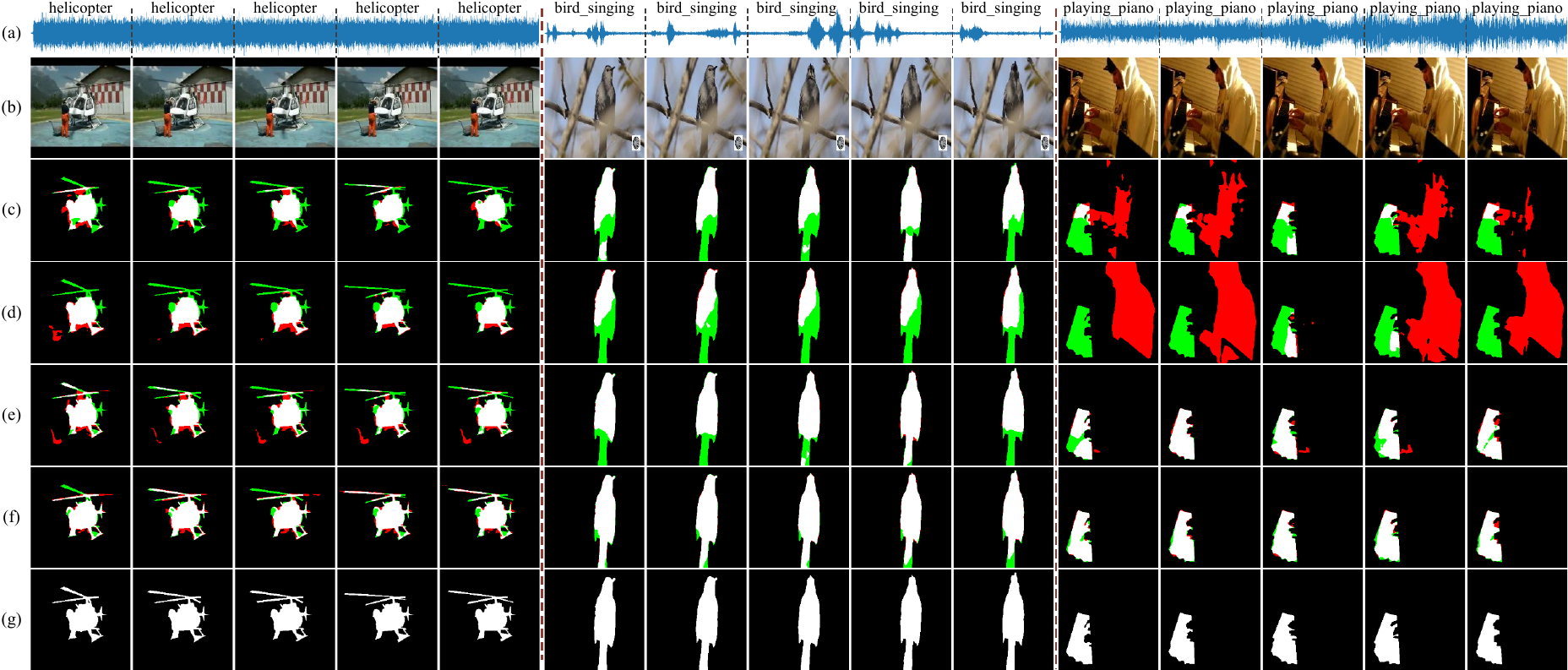}
  \caption{Qualitative comparisons on S4. (a) audio, (b) frames, (c) AVSBench \cite{17}, (d) AVSegFormer \cite{21}, (e) COMBO \cite{39}, (f) DGCM-AVS (Ours), (g) GT. The green denotes missed regions and red indicates over-segmented areas relative to GT.}
  \label{fig:4}
\end{figure*}

\section{Experiments}
\subsection{Dataset}

AVSBench-Object \cite{16} consists of two subsets: Single Sound Source Segmentation (S4) and Multiple Sound Source Segmentation (MS3). The S4 subset includes 4,932 five-second video clips, each containing a single sounding object. Only the first frame of each five-frame sequence is annotated, yielding a semi-supervised setting. The dataset uses a 70/15/15 train/val/test split. The MS3 subset contains 424 fully annotated videos with multiple simultaneous sound sources, reflecting more complex auditory scenes. 

AVSBench-Semantic (AVSS) \cite{17} extends AVSBench-Object to the audio-visual semantic segmentation task. It follows the same train/val/test split as S4. The dataset includes all AVSBench-Object samples and adds semantic labels to them. It also adds new 10-second videos with frame-level semantic masks, bringing the total to 11,356 videos. This extension increases task complexity and makes semantic modeling and generalization more challenging.

\vspace{-0.2cm} 
\subsection{Implementation Details}
\label{4.2}

\noindent\textbf{Training settings. }All experiments are conducted on NVIDIA RTX 4090 GPU with input resolution fixed at \(224 \times 224\). We use Adam with an initial learning rate of \(10^{-4}\), weight decay of 0.05, and a batch size of 6. The model is trained for 90K iterations on S4 and AVSS, and 20K on MS3.

\noindent\textbf{Metrics. }Following the standard protocol \cite{17,21}, we use the Jaccard index \cite{everingham2010pascal} (\( \mathcal{M}_\mathcal{J} \)) and the F-score (\( \mathcal{M}_\mathcal{F} \)) as evaluation metrics. The Jaccard index is defined as \(\mathcal{M}_{\mathcal{J}}=\frac{\left| P\cap G \right|}{\left| P\cup G \right|}\), where \(P\) and \(G\) denote the predicted and ground-truth masks, and \(\left| \cdot \right|\) denotes the number of pixels. The F-score is defined as \(\mathcal{M}_{\mathcal{F}}=\frac{\left( 1+\beta ^2 \right) \times Precision\times Recall}{\beta ^2\times Precision+Recall}\), where \(\beta ^2\) is set to 0.3. Precision measures how accurate the predicted foreground pixels are, and Recall measures how completely the true foreground pixels are detected. \( \mathcal{M}_\mathcal{J} \) measures the region-level intersection-over-union between the predicted and ground-truth masks, and reflects how well the overall shape and coverage of the foreground region are matched. \( \mathcal{M}_\mathcal{F} \) depends on both Precision and Recall, and focuses more on the quality of the segmentation boundaries, being sensitive to false positives and false negatives along object edges. These two metrics evaluate the audio-visual segmentation results from the complementary perspectives of boundary accuracy and region similarity, and thus provide a more comprehensive assessment of the model performance.

\noindent\textbf{Loss. }The total loss includes classification and segmentation terms, defined as \( \mathcal{L}=\lambda_{bce}\mathcal{L}_{bce}+\lambda_{dice}\mathcal{L}_{dice}+\lambda_{ce}\mathcal{L}_{ce} \), where \(\mathcal{L}_{ce}\) is the cross-entropy loss for class prediction, and \(\mathcal{L}_{bce}\) and \(\mathcal{L}_{dice}\) are the binary cross-entropy loss and Dice loss used for mask supervision. Following prior works \cite{11,39,40}, the weights are set to \( \lambda_{bce} = 5 \), \( \lambda_{dice} = 5 \), and \( \lambda_{ce} = 2 \).

\vspace{-0.3cm} 
\subsection{Main Results}

\noindent\textbf{Quantitative Analysis.} Tab. \ref{tab:1} summarizes the quantitative performance of DGCM-AVS on the S4, MS3, and AVSS benchmarks, compared with representative methods.
On the S4 benchmark, DGCM-AVS-R50 outperforms prior methods by 1.6 \( \mathcal{M}_\mathcal{J} \) and 1.2 \( \mathcal{M}_\mathcal{F} \), while DGCM-AVS-PVT achieves gains of 0.5 \( \mathcal{M}_\mathcal{J} \) and 0.5 \( \mathcal{M}_\mathcal{F} \). 
On MS3, DGCM-AVS-R50 achieves substantial improvements of 1.6 \( \mathcal{M}_\mathcal{J} \) and 7.8 \( \mathcal{M}_\mathcal{F} \), and DGCM-AVS-PVT improves by 1.2 \( \mathcal{M}_\mathcal{J} \) and 2.5 \( \mathcal{M}_\mathcal{F} \). 
On AVSS, our model also attains the best performance: DGCM-AVS-R50 improves by 6.3 \( \mathcal{M}_\mathcal{J} \) and 6.2 \( \mathcal{M}_\mathcal{F} \), and DGCM-AVS-PVT achieves 4.3 \( \mathcal{M}_\mathcal{J} \) and 4.1 \( \mathcal{M}_\mathcal{F} \). 

Overall, DGCM-AVS consistently surpasses prior methods across all benchmarks, with particularly notable improvements on the more complex MS3 and AVSS datasets. While the gain on S4 is relatively small, this can be attributed to its simple scenes and limited interference, where most methods already perform strongly. In contrast, the significant improvements on MS3 and AVSS highlight that depth cues are especially beneficial in challenging scenarios involving multiple sound sources, occlusion, or cluttered backgrounds. These results support that depth can act as an effective bridge for cross-modal alignment and provide complementary structural information that is difficult to obtain from audio-visual cues alone.

\noindent\textbf{Qualitative Analysis on S4.} Fig. \ref{fig:4} presents additional qualitative comparisons on the S4. In the first set of examples, the red tail of the helicopter closely resembles the background factory in color and texture, making boundary segmentation highly challenging. Methods such as AVSBench and AVSegFormer, which do not incorporate explicit spatial cues, struggle to separate the object from the background, while COMBO provides only limited improvement due to imperfect visual prompts. In contrast, DGCM-AVS leverages depth cues via the DADM to produce sharper boundaries and maintain intra-object consistency. Furthermore, DGPF treats depth as a bridging modality and employs a two-stage strategy to facilitate more explicit alignment between audio cues and visual targets.

In the second set of examples, the sounding object is partially occluded by tree branches, and its tail shares textures with the surroundings. All baseline methods degrade noticeably under occlusion, whereas DGCM-AVS demonstrates more stable segmentation, even though minor edge inaccuracies remain. In the third set of examples, AVSBench and AVSegFormer show obvious target recognition errors, while DGCM-AVS effectively leverages geometric cues from depth information to preserve the structural integrity of the object and achieve higher segmentation accuracy than COMBO. 

On the S4, although the scenes often involve challenges such as color similarity, texture confusion, or partial occlusion, DGCM-AVS effectively integrates audio-visual features with spatial structural cues to preserve object integrity and sharpen boundaries, thereby achieving higher segmentation accuracy than existing methods.

\begin{figure*}
  \centering
  \includegraphics[width=1\linewidth]{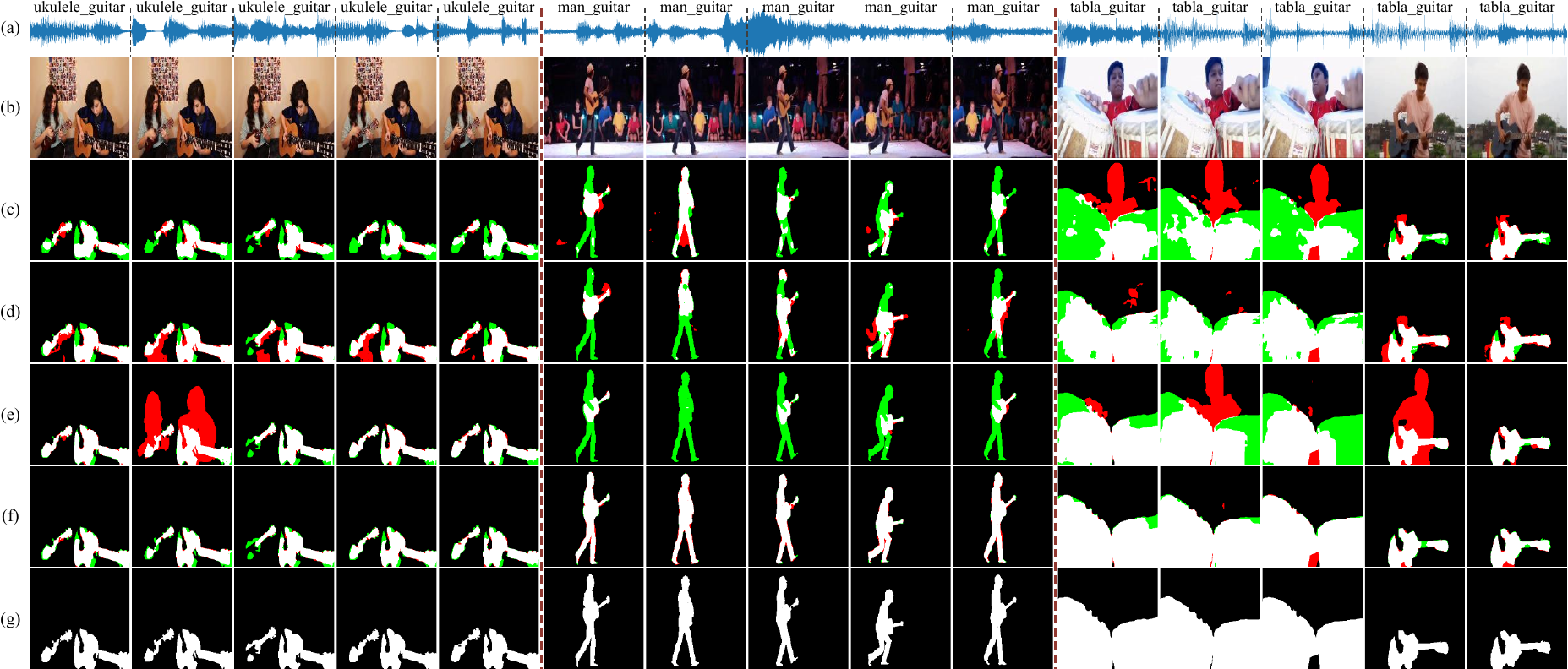}
  \caption{Qualitative comparisons on MS3. (a) audio, (b) frames, (c) AVSBench \cite{17}, (d) AVSegFormer \cite{21}, (e) COMBO \cite{39}, (f) DGCM-AVS (Ours), (g) GT. The green denotes missed regions and red indicates over-segmented areas relative to GT.}
  \label{fig:5}
  \vspace{-0.4cm}
\end{figure*}

\noindent\textbf{Qualitative Analysis on MS3.} Fig. \ref{fig:5} shows more comparison results on MS3. In the first set of examples, AVSBench and AVSegFormer struggle to capture the boundaries of the instrument accurately. This is mainly because the sounding objects are surrounded by high-frequency visual patterns, which interfere with the true contours and lead to incorrect predictions. COMBO shows instability, mistakenly identifying the person in the scene as a sound source in some frames.

The second set of examples includes more semantically similar distractors in the background. Under this challenging condition, our method produces results that are most aligned with the ground truth and successfully identifies all sounding objects. In contrast, AVSBench and AVSegFormer are clearly affected by background noise and mistakenly include silent regions in their predictions. COMBO fails to segment the “man” object, mainly because it relies on prompt information from SAM. Such prompts may break the integrity of the target object, leading to incomplete predictions. In the third set of examples, the video contains different sounding objects at different times, resulting in temporal discontinuity of the targets. In this case, our method shows more stable performance than the baselines and accurately segments each sounding object.

These results on MS3 suggest that depth, as a carrier of spatial structure, enhances the model’s understanding of complex scenes and helps distinguish object boundaries more effectively. Moreover, aligning audio and visual features through depth promotes more effective cross-modal interaction, guiding the model toward potential sounding regions and improving segmentation quality.

\begin{figure*}
  \centering
  \includegraphics[width=1\linewidth]{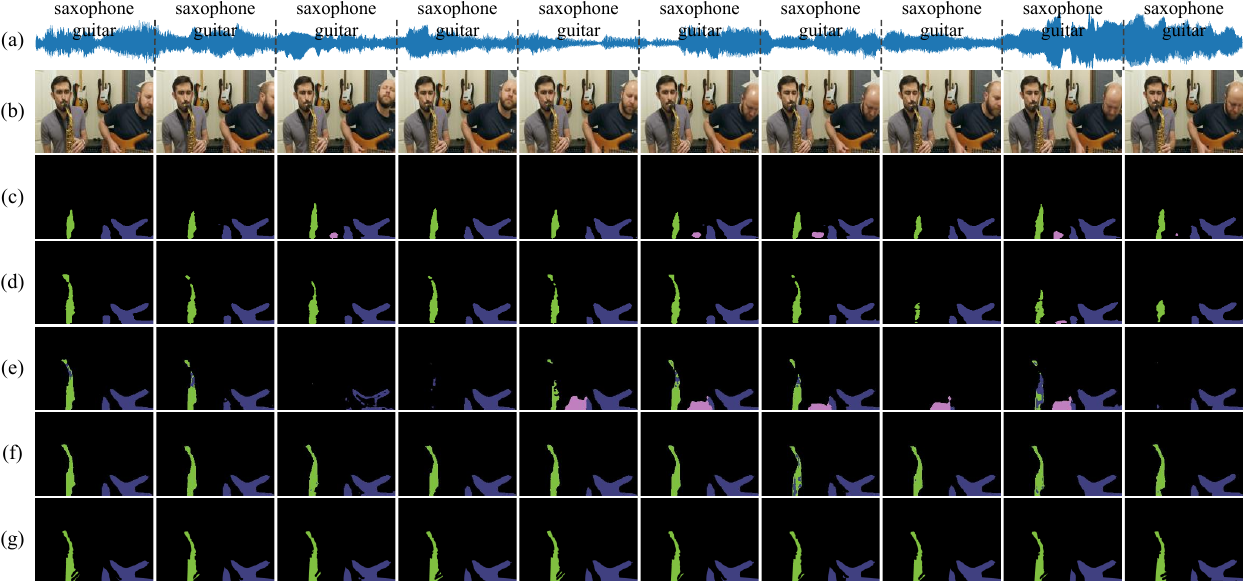}
  \caption{Qualitative comparisons on AVSS. We provide a full visualization of the semantic segmentation performed on 10-frame sequences. The letter labels represent, in order: (a) audio, (b) frames, (c) AVSBench \cite{17}, (d) AVSegFormer \cite{21}, (e) COMBO \cite{39}, (f) DGCM-AVS (Ours), (g) GT. }
  \label{fig:6}
\end{figure*}

\noindent\textbf{Qualitative Analysis on AVSS.} The AVSS dataset features higher sample complexity and longer video sequences. To illustrate the model’s performance under such challenging conditions, we present qualitative results on a representative 10-frame sample, as shown in Fig. \ref{fig:6}. In terms of segmentation completeness, AVSBench and AVSegFormer produce coarse boundary delineation for sounding objects, leading to reduced structural integrity. For target identification, COMBO exhibits missing sounding objects and misclassification errors, and its predictions show temporal inconsistency across frames. In contrast, our method achieves the best semantic segmentation results. This advantage comes from two main aspects. First, depth information provides a clearer structure of the scene. Our proposed DADM module fully leverages this geometric information to help maintain feature consistency within objects. Second, DGPF maps audio and visual features into a feature space guided by depth, which enhances the alignment between modalities. These results validate the effectiveness and feasibility of using depth information in AVS.

\begin{table*}[t]
    \renewcommand\arraystretch{1.1} 
    \small
    \centering
    \caption{Efficiency and Performance Comparison on S4. Bold indicates the best result.}
    \setlength{\tabcolsep}{3.2mm} 
    {
        \begin{tabular}{lccccccc}
        \toprule
        Method                & Audio Backbone & Backbone  & \(\mathcal{M}_\mathcal{J}\) (S4)  & \(\mathcal{M}_\mathcal{F}\) (S4) & Params  & FLOPs  & Inference Time \\
        \midrule
        AVSBench \cite{17}    & VGGish   & ResNet-50   & 72.8           & 84.8    & 90.7M      & 166.6G                & 17.7ms    \\
        AVSegFormer \cite{21} & VGGish   & ResNet-50   & 76.4           & 86.7    & 150.9M      & 210.1G                 & 176.5ms        \\       
        DGCM-AVS (Ours)        & VGGish   & ResNet-50   & \textbf{83.3}  & \textbf{91.3}     & 224.3M      & 183.4G        & 60.7ms \\
        \bottomrule
        \end{tabular}
    }
    \label{tab:2}
    \vspace{-0.4cm}
\end{table*}

\noindent\textbf{Efficiency and Performance Comparison. }Tab. \ref{tab:2} compares AVSBench, AVSegFormer, and our DGCM-AVS on the S4 benchmark. For a fair comparison, all three methods use VGGish as the audio backbone and ResNet-50 as the visual backbone. In terms of performance, DGCM-AVS achieves the best results, with \(\mathcal{M}_\mathcal{J}\) of 83.3 and \(\mathcal{M}_\mathcal{F}\) of 91.3, significantly outperforming the two representative methods. This gain mainly stems from our proposed depth-guided collaborative modeling mechanism, which enables the model to more stably constrain cross-modal correspondence under candidate confusion, occlusions, and cluttered backgrounds, thereby reducing attention drift and boundary leakage and improving both localization and boundary quality. In contrast, AVSBench adopts a lightweight convolution-based modeling paradigm and thus runs faster, but its cross-modal interaction and structural constraints are relatively limited in complex scenes, leading to inferior accuracy. AVSegFormer benefits from stronger global modeling and achieves improved accuracy, but it also introduces higher computational overhead, resulting in the slowest inference. From the perspective of computational complexity, DGCM-AVS requires 183.4G FLOPs, which is notably lower than AVSegFormer (210.1G), indicating that our performance improvement is not simply obtained by substantially increasing computation. Regarding inference time, DGCM-AVS runs at 60.7ms, falling between AVSBench and AVSegFormer, and is considerably faster than AVSegFormer. Although DGCM-AVS incorporates depth information and thus has a larger parameter scale, it still follows a single forward-pass pipeline without additional iterative refinement, maintaining a reasonable inference cost while achieving higher accuracy. In practical deployments, parameter storage is typically cheaper than sustained computation, and parameters mainly affect the static memory footprint rather than per-inference computation. Considering the advantages in both accuracy and efficiency, DGCM-AVS demonstrates a clear competitive edge.

\begin{figure}
  \centering
  \includegraphics[width=1\linewidth]{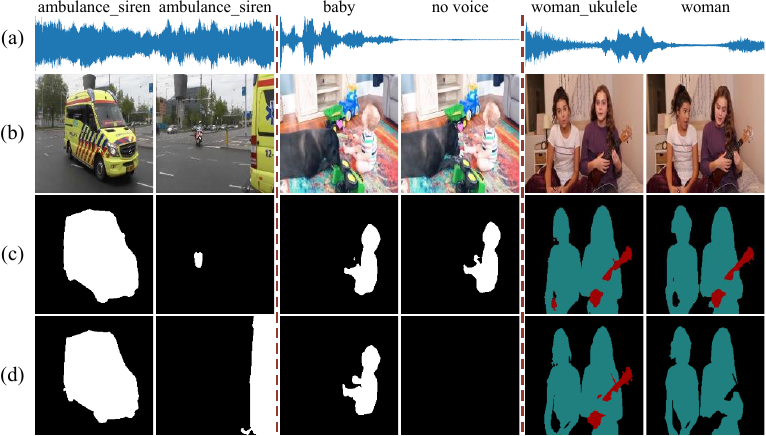}
  \caption{Failure cases of DGCM-AVS on S4, MS3, and AVSS (left to right). For each example, we show (a) audio, (b) frames, (c) DGCM-AVS (ours), and (d) GT.}
  \label{fig:fail}
\end{figure}

\noindent\textbf{Failure Case Analysis.} Although DGCM-AVS achieves strong performance on the three benchmarks, it still shows limitations in a few complex audio-visual scenes. Fig. \ref{fig:fail} presents three representative challenging cases from S4, MS3, and AVSS. In the first case, only the ambulance is sounding, but in the second frame the ambulance almost moves out of view and a police motorbike appears. Because the two sirens sound very similar, the model wrongly localizes the motorbike as the sounding object. In the second case, the baby makes sound only in the first frame and is correctly segmented there, but in the fully silent second frame the model still follows the previous prediction and labels the baby as sounding. The third case shows a similar error when the ukulele becomes silent but is still predicted as a sounding region.

These results indicate that DGCM-AVS, while stable in capturing object contours with the help of depth, is not robust enough when the scene contains multiple similar sound sources, sources moving out of view, or rapid changes in the sounding state. We believe there are two main reasons. First, such complex patterns are very rare in current benchmarks, so the training data under-represents these cases. Second, the continuous and aligned audio–video inputs cause strong temporal dependence, or even entanglement, in the audio features, which makes the model biased toward historical objects on silent frames or frames with source switching.

\subsection{Ablation Study}

\begin{figure*}[h]
  \centering
  \includegraphics[width=1\linewidth]{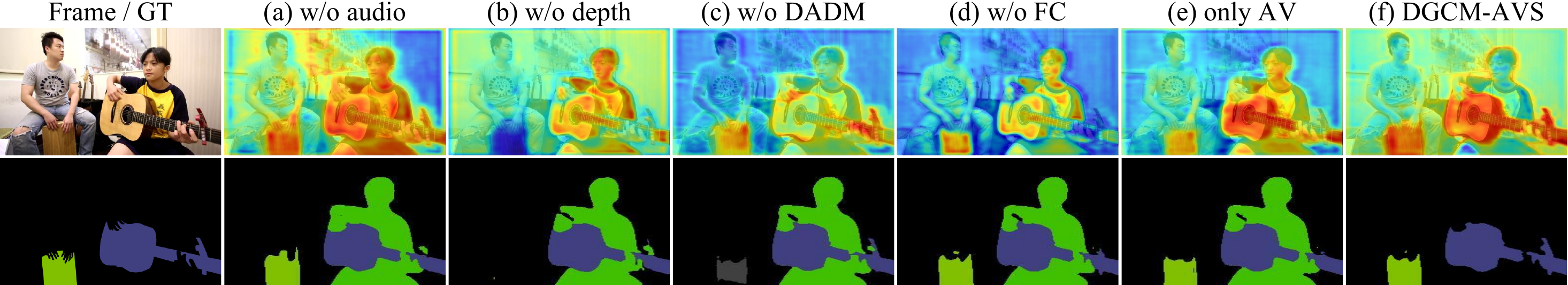}
  \caption{Visualization of the ablation study to show the efficacy of each proposed component.}
  \label{fig:7}
  \vspace{-0.4cm}
\end{figure*}

\noindent\textbf{Validation of DADM and Backbone Weights.} As shown in Tab. \ref{tab:3}, \textit{shared encoder} refers to using the same encoder for both visual and depth information. \textit{DGCM-AVS} uses independent encoders for each modality. \textit{w/o depth} indicates that depth information is excluded from the model, and both branches in DADM operate solely on visual features. \textit{w/o DADM} removes the entire module, and depth and visual features are fused by direct addition. The results show that using independent encoders helps the model better capture modality-specific structures, leading to improved performance. Introducing depth information helps the model better localize and segment sounding objects by providing additional spatial cues. Adding the DADM enables complementary fusion of low-frequency visual features and high-frequency depth cues, improving edge sensitivity and robustness to noise, which leads to more accurate segmentation.

\begin{table}[t]
    \vspace{-0.4cm}
    \renewcommand\arraystretch{1.1} 
    \small
    \centering
    \caption{Validation of DADM and Weights on ResNet-50.}
    \setlength{\tabcolsep}{4mm} 
    {
        \begin{tabular}{lcccc}
        \toprule
        \multirow{2.5}{*}{Method} & \multicolumn{2}{c}{S4} & \multicolumn{2}{c}{AVSS} \\ \cmidrule(r){2-3} \cmidrule(r){4-5}
                                & \(\mathcal{M}_\mathcal{J}\)        & \(\mathcal{M}_\mathcal{F}\)        & \(\mathcal{M}_\mathcal{J}\)        & \(\mathcal{M}_\mathcal{F}\)         \\ \midrule
        shared encoder          & 82.9       & 91.0      & 39.4        & 43.3       \\
        w/o depth               & 81.1       & 90.0      & 37.0        & 40.9       \\
        w/o DADM                & 83.2       & 90.9      & 38.7        & 42.4       \\
        DGCM-AVS                & \textbf{83.3}       & \textbf{91.3}      & \textbf{39.6}        & \textbf{43.5}       \\
        \bottomrule
        \end{tabular}
    }
    \label{tab:3}
    \vspace{-0.4cm}
\end{table}

\begin{table}[t]
    \renewcommand\arraystretch{1.1} 
    \small
    \centering
    \caption{Validation the \(k_h\) and \(k_l\) window size of DADM.}
    \setlength{\tabcolsep}{3.7mm} 
    {
        \begin{tabular}{lcccc}
        \toprule
        \multirow{2.5}{*}{Method} & \multicolumn{2}{c}{S4} & \multicolumn{2}{c}{AVSS} \\ \cmidrule(r){2-3} \cmidrule(r){4-5}
                                & \(\mathcal{M}_\mathcal{J}\)        & \(\mathcal{M}_\mathcal{F}\)        & \(\mathcal{M}_\mathcal{J}\)        & \(\mathcal{M}_\mathcal{F}\)         \\ \midrule
        \(\text{DADM}_{k_l=7,k_h=5}\)          & 83.1       & 91.1      & 39.2        & 43.1       \\
        \(\text{DADM}_{k_l=7,k_h=3}\)               & 82.7       & 90.8      & 38.9        & 42.6       \\
        \(\text{DADM}_{k_l=5,k_h=3}\)                & \textbf{83.3}       & \textbf{91.3}      & \textbf{39.6}        & \textbf{43.5}       \\
        \bottomrule
        \end{tabular}
    }
    \label{tab:4}
\end{table}

\noindent\textbf{Validation the Window Size of DADM.} As shown in Tab.~\ref{tab:4}, we evaluate different window sizes \((k_l, k_h)\) used in the low-frequency aggregation branch and the high-frequency residual branch of DADM. The best performance is achieved with \((k_l = 5, k_h = 3)\). Specifically, \(k_l = 5\) provides sufficient context to produce a smoother and more stable aggregated representation, which helps preserve region-level semantic consistency and supports robust alignment with audio cues. In contrast, a smaller local window \(k_h = 3\) better captures rapid spatial changes around object boundaries, making the explicitly modeled residual cue \(\mathcal{F}_d-\mathcal{F}_d'\) more concentrated and thus improving boundary sensitivity. Larger window settings \(k_l=7\) lead to over-smoothing and neighborhood mixing, which blur fine boundary details and increase computational cost. Overall, we adopt \((k_l=5, k_h=3)\) as it achieves the best performance while preserving stable region responses and clear boundaries.

\begin{table}[t]
\vspace{-0.4cm}
    \renewcommand\arraystretch{1.1} 
    \small
    \centering
    \caption{Validation of DGPF structures on ResNet-50.}
    \setlength{\tabcolsep}{4.5mm} 
    {
        \begin{tabular}{lcccc}
        \toprule
        \multirow{2.5}{*}{Method}  & \multicolumn{2}{c}{S4} & \multicolumn{2}{c}{AVSS} \\ \cmidrule(r){2-3} \cmidrule(r){4-5}
                                 & \(\mathcal{M}_\mathcal{J}\)        & \(\mathcal{M}_\mathcal{F}\)        & \(\mathcal{M}_\mathcal{J}\)         & \(\mathcal{M}_\mathcal{F}\)         \\ \midrule
        only AV      & 82.9       & 90.9      & 38.7        & 42.4       \\
        w/o audio      & 82.4       & 91.1      & 38.0        & 41.8       \\
        w/o FC       & 82.9       & 91.2      & 39.2        & 43.0       \\
        DGCM-AVS      & \textbf{83.3}       & \textbf{91.3}      & \textbf{39.6}        & \textbf{43.5}       \\
        \bottomrule
    \end{tabular}
    }
    \label{tab:5}
    \vspace{-0.4cm}
\end{table}

\noindent\textbf{Validation of DGPF.} As shown in Tab. \ref{tab:5}, \textit{only AV} refers to directly performing the Target Search stage for audio-visual alignment, without involving depth as an intermediate bridge.
\textit{w/o audio} removes the audio modality completely, keeping only the direct fusion of visual and depth information. \textit{w/o FC} removes the Feature Calibrator, which reduces the model’s focus on potential target regions. Compared to these variants, the full DGPF model demonstrates the effectiveness of each component and the benefit of their integration. The Feature Calibrator enhances the model’s attention to potential targets within visual and depth features, resulting in more pronounced improvements on the AVSS dataset. Audio features further serve as guidance, enabling the model to locate sounding objects based on auditory cues. Built upon these cues, DGPF adopts a two-stage strategy in which depth acts as a bridge between modalities, progressively aligning audio and visual features to support accurate segmentation.

\begin{table}[t]
    \renewcommand\arraystretch{1.1} 
    \small
    \centering
    \caption{The impact of different depth information.}
    \setlength{\tabcolsep}{4.5mm} 
    {
        \begin{tabular}{lcccc}
        \toprule
        \multirow{2.5}{*}{Method}  & \multicolumn{2}{c}{S4} & \multicolumn{2}{c}{AVSS} \\
        \cmidrule(r){2-3} \cmidrule(r){4-5}
            & \(\mathcal{M}_\mathcal{J}\) & \(\mathcal{M}_\mathcal{F}\)     & \(\mathcal{M}_\mathcal{J}\) & \(\mathcal{M}_\mathcal{F}\)     \\ \midrule
        + DP           & 82.0 & 90.5   & 37.7 & 41.7   \\
        + DA           & 83.2 & 90.9   & 37.6 & 41.7   \\
        + DA-V2        & \textbf{83.3} & \textbf{91.3} & \textbf{39.6} & \textbf{43.5} \\
        \bottomrule
        \end{tabular}
    }
    \label{tab:6}
\end{table}

\noindent\textbf{Visualization for Ablation Studies. }Fig. \ref{fig:7} illustrates the visual impact of different ablation settings. It can be observed that removing any key component leads to distinct degradations. Without audio guidance, (a) the model loses the constraint of sounding cues, and the attention is more easily attracted by visually salient regions and diffuses into the background, resulting in evident false positives. Without depth information, (b) the lack of geometric structure and boundary cues weakens the model’s ability to characterize object shape, making predictions prone to adhesion to adjacent regions and exhibiting both false positives and false negatives. Furthermore, removing DADM (Fig. \ref{fig:7}(c)) noticeably reduces the model’s perception of scene structural information, while intra-object features become less continuous and eventually cause false detections, indicating that DADM helps maintain intra-object consistency and improves boundary sensitivity between adjacent objects. Removing FC (Fig. \ref{fig:7}(d)) makes the fusion and calibration process less stable, with weaker and less complete attention responses, which tends to introduce noisy regions and miss thin structures. When only audio-visual fusion is retained (Fig. \ref{fig:7}(e)), although the attention can still fall near the sounding objects, the absence of structural constraints makes the focus more likely to drift, leading to coarse boundaries and unstable object recognition. In contrast, the full DGCM-AVS model (Fig. \ref{fig:7}(f)) can more stably focus on the sounding objects, producing predictions with region coverage more consistent with the GT and clearer objects contours. These results suggest that audio provides a semantic anchor for sounding objects, depth offers geometric and boundary constraints, and DADM and DGPF work complementarily to achieve more robust and accurate audio-guided segmentation.

\begin{figure}
\vspace{-0.4cm} 
  \centering
  \includegraphics[width=1\linewidth]{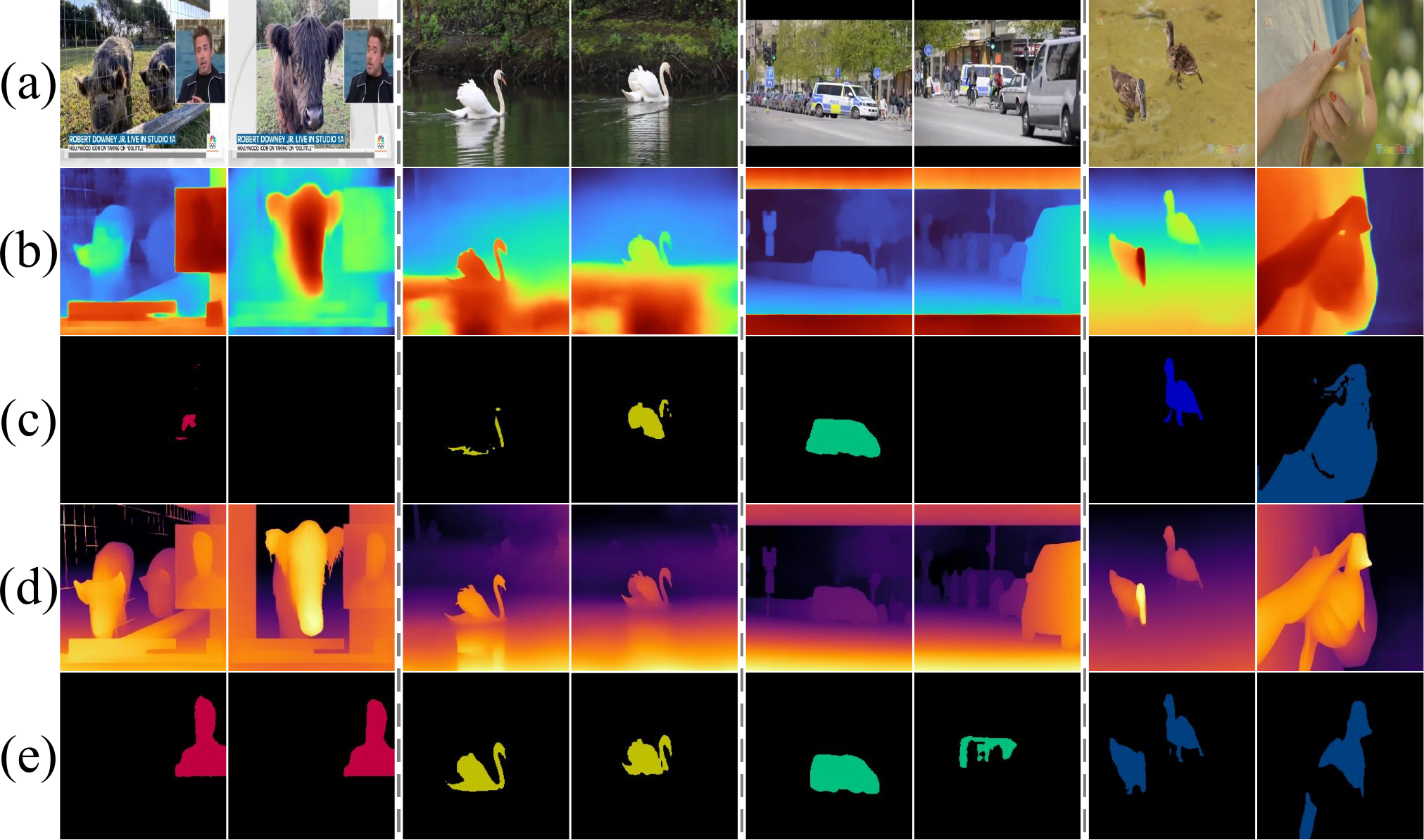}
  \caption{The impact of depth information. (a) frames, (b) DP results, (c) AVS results based on DP , (d)  DA-V2 results, (e) AVS results based on DA-V2.}
  \label{fig:8}
  \vspace{-0.4cm}
\end{figure}

\noindent\textbf{Validation of the Impact of Different Depth Information.} As shown in Tab. \ref{tab:6}, we compare three recent models used as depth generators to study how different types of depth information can help align audio and visual modalities. Due to the high resource demands of the three depth generators with their best pretrained weights, we adopt a two-stage pipeline: depth information is first extracted, followed by the AVS task.
Depth Pro (DP) \cite{61} is based on geometry-aware modeling, while Depth Anything (DA) \cite{34} and Depth Anything v2 (DA-V2) \cite{35} rely on vision-centric pretraining. Among them, DA-V2 achieves clearly better results than the others. As in Fig. \ref{fig:8}, we compare the robustness of DP and DA-V2 in generating depth information under different scenes. DA-V2 produces more stable and accurate depth information than DP, especially in complex environments. The results show that although high-quality depth information can further improve performance, it still requires the proposed DGCM-AVS framework to effectively exploit such spatial cues in order to realize their full potential. Fortunately, with the recent advances in depth estimation, it is now possible to obtain more reliable depth, which provides strong support for further research in audio-visual segmentation.

\section{Conclusion}
In this work, we propose DGCM-AVS, a depth-guided framework for audio-visual segmentation. To the best of our knowledge, this is the first AVS framework that explicitly uses depth as a bridging cue to guide alignment between audio and visual modalities. The Depth-Aware Dynamic Modulator effectively leverages boundary-sensitive high-frequency cues from depth maps and semantically informative low-frequency cues from visual features. It improves intra-object feature consistency while enhancing inter-object discrimination. The Depth-Guided Progressive Fusion adopts a two-stage design that uses depth as a bridge to progressively align audio cues with visual features. DGCM-AVS achieves promising results on both AVSBench-Object and AVSBench-Semantic benchmarks. These results suggest that incorporating depth cues is a promising direction for improving robustness in AVS.

\section{Future Work}

Future work could address the observed failure modes from both data and methodological perspectives. On the data side, it would be beneficial to incorporate more training videos with complex audio-visual scenes to improve generalization to long-tail cases. On the methodological side, it would be useful to explore explicit temporal decoupling and contrastive learning strategies that encourage models to focus on sound cues synchronized with the current frame and suppress residual signals from previous frames. In addition, it would be interesting to study AVS in embodied settings, where segmentation can be tightly coupled with perception--action loops to support more reliable scene understanding and interaction in real-world environments.


\bibliographystyle{IEEEtran}
\bibliography{ref}

\vfill

\end{document}